\documentclass[remotesensing,article,accept,pdftex,moreauthors]{Definitions/mdpi} 

\firstpage{1} 
\pubvolume{18}
\issuenum{6}
\articlenumber{915}
\pubyear{2026}
\copyrightyear{2026}
\externaleditor{Francisco Javier Mesas Carrascosa and Fernando Pérez Porras} 
\datereceived{19 February 2026} 
\daterevised{13 March 2026} 
\dateaccepted{16 March 2026} 
\datepublished{17 March 2026} 

\usepackage{amsmath,amssymb,amsfonts}
\usepackage{algorithmic}
\usepackage{graphicx}
\usepackage{textcomp}
\usepackage{xcolor}
\usepackage{longtable} 
\usepackage{epsfig}
\usepackage{graphicx}
\usepackage{amsmath,amssymb}
\usepackage{booktabs}
\usepackage[accsupp]{axessibility} 
\usepackage{multirow} 

\Title{{BAWSeg:} A UAV Multispectral Benchmark for Barley Weed~Segmentation}

\Author{%
{Haitian~Wang} 
$^{1}$\orcidA{}, Xinyu~Wang $^{1}$\orcidB{}, Muhammad~Ibrahim $^{2}$\orcidC{}, Dustin~Severtson $^{2}$\orcidD{} and {Ajmal~Mian} 
 $^{1,}$*\orcidE{}%
}

\AuthorNames{Haitian Wang; Muhammad Ibrahim; Dustin Severtson; Xinyu Wang; Ajmal Mian}

\address{%
$^{1}$ \quad Department of Computer Science and Software Engineering, The University of Western Australia, 35 Stirling Highway, Crawley, WA 6009, Australia; 
haitian.wang@uwa.edu.au (H.W.); xinyu.wang@uwa.edu.au (X.W.)\\
$^{2}$ \quad Department of Primary Industries and Regional Development (DPIRD), Government of Western Australia, \linebreak 1 Nash {St.,} 
 Perth, WA 6000, Australia; 
muhammad.ibrahim@dpird.wa.gov.au (M.I.); dustin.severtson@dpird.wa.gov.au (D.S.)
}

\corres{Correspondence: ajmal.mian@uwa.edu.au}

\abstract{
{Accurate}
 weed mapping in cereal fields requires pixel-level segmentation from unmanned aerial vehicle (UAV) imagery that remains reliable across fields, seasons, and illumination. Existing multispectral pipelines often depend on thresholded vegetation indices, which are brittle under radiometric drift and mixed crop--weed pixels, or on single-stream convolutional neural network (CNN) and Transformer backbones that ingest stacked bands and indices, where radiance cues and normalized index cues interfere and reduce sensitivity to small weed clusters embedded in crop canopy. We propose VISA (Vegetation Index and Spectral Attention), a two-stream segmentation network that decouples these cues and fuses them at native resolution. {The radiance stream learns from calibrated five-band reflectance using local residual convolutions, channel recalibration, spatial gating, and skip-connected decoding, which preserve fine textures, row boundaries, and small weed structures that are often weakened after ratio-based index compression.
} The index stream operates on vegetation-index maps with windowed self-attention to model local structure efficiently, state-space layers to propagate field-scale context without quadratic attention cost, and Slot Attention to form stable region descriptors that improve discrimination of sparse weeds under canopy mixing. To support supervised training and deployment-oriented evaluation, we introduce BAWSeg, a four-year UAV multispectral dataset collected over commercial barley paddocks in Western Australia, providing radiometrically calibrated blue, green, red, red edge, and near-infrared orthomosaics, derived vegetation indices, and dense crop, weed, and other labels with leakage-free block splits. On BAWSeg, VISA achieves 75.6\% mean Intersection over Union (mIoU) and 63.5\% weed Intersection over Union (IoU) with 22.8 M parameters, outperforming a multispectral SegFormer-B1 baseline by 1.2 mIoU and 1.9 weed IoU. Under cross-plot and cross-year protocols, VISA maintains 71.2\% and 69.2\% mIoU, respectively. {The full BAWSeg benchmark dataset, VISA code, trained model weights, and protocol files will be released upon publication.}
}

\keyword{UAV; multispectral; weed mapping; barley; semantic segmentation; vegetation indices; radiometric calibration; orthomosaic; domain shift; benchmark dataset}

\newcommand{\CIMain}{0.004}
\newcommand{\CIAbl}{0.007}
\newcommand{\ParamCountM}{22.8}

\begin{document}

\section{Introduction}

Barley is a major cereal crop cultivated across temperate and semi-arid regions worldwide, forming a core component of cereal-based farming systems alongside wheat and canola \cite{abares2023crop,giwa2023barley}. Across these systems, weeds remain the primary biotic constraint on yield, causing around 13\% average global yield loss and billions of dollars of annual economic impact due to competition and herbicide resistance \cite{pacanoski2021weeds,oesk2021weedloss,llewellyn2016impact,powles2010resistance}. Herbicide-resistant populations of grass and broadleaf weeds are now common in many intensive cereal regions, motivating strong demand for spatially explicit weed management strategies.

In barley-based cropping systems, species such as annual ryegrass (\textit{Lolium rigidum}) and wild radish (\textit{Raphanus raphanistrum}) often dominate the weed spectrum, driving double-digit yield and quality losses when control is delayed or ineffective \cite{walsh2012harvest,broster2019survey,reeves2018ryegrass,hashem2020wildradish}. These pressures highlight the need for high-resolution sensing and mapping approaches capable of delineating weed patches at sub-field scales to support site-specific management \cite{chlingaryan2018uav,lambert2018weedmaps}.

Current practice in precision weed management combines manual scouting with proximal and remote sensing technologies, including satellite imagery, crewed aircraft surveys, and UAV-based multispectral imaging. These pipelines typically compute vegetation indices and apply threshold rules or shallow classifiers to distinguish crop, weed, and soil classes \cite{lopezgranados2011weedreview,chlingaryan2018uav,lopezgranados2016sunflower}. However, their reliance on hand-crafted indices, per-tile parameter tuning, and independent pixel or patch decisions limits robustness under variable illumination, canopy closure, and mixed crop--weed cover \cite{perezortiz2016patterns,jin2023corn,li2023weedreview}. Deep convolutional and Transformer-based encoder--decoder architectures improve spatial coherence but often rely on a single concatenated feature stream and are rarely evaluated under strong spatial or temporal domain shifts in cereal systems \cite{ronneberger2015unet,badrinarayanan2017segnet,chen2018deeplabv3plus,lottes2017icra}. Despite rapid progress, publicly available benchmarks for weed segmentation do not yet match the practical deployment setting of cereal paddocks~\cite{wang2024city}. Many widely used weed datasets focus on image-level recognition or close-range field imagery, rather than field-scale UAV orthomosaics with pixel-level crop--weed masks. Large-scale aerial agriculture datasets (e.g., Agriculture-Vision) provide RGB--NIR patches with anomaly-style masks such as weed clusters, but they are not designed as calibrated multi-band UAV orthomosaics for explicit crop--weed semantic segmentation across multiple seasons and fields. As a consequence, prior work often reports results on single-field or single-season splits, making it difficult to quantify generalization under realistic spatial and temporal shifts.

To address these limitations, we propose VISA (Vegetation Index and Spectral Attention), a two-stream segmentation network that separates radiance-based and index-based reasoning. The radiance branch preserves fine row and patch boundaries through residual attention modules, while the index branch captures plot-level contextual patterns using attention and state-space mechanisms. Native-resolution features from both streams are fused to produce accurate crop--weed--background predictions, demonstrating improved segmentation performance under mixed canopy, variable illumination, and domain-shift~conditions.

\begin{figure}[H]
\includegraphics[width=\linewidth]{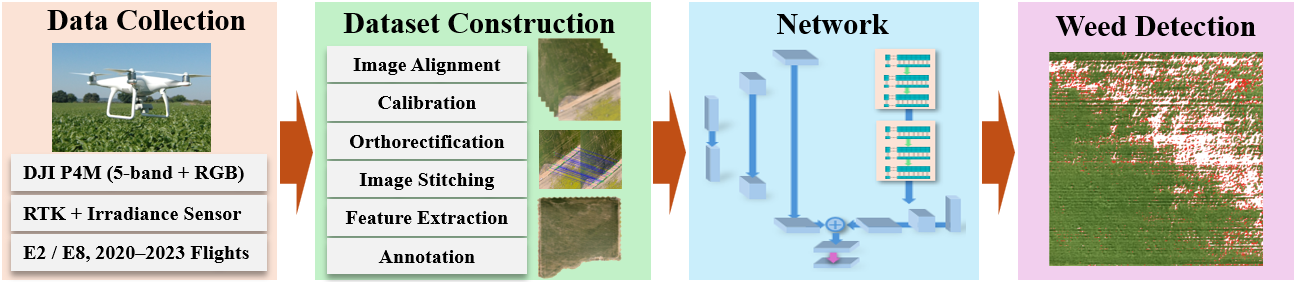}
\caption{{Overview} 
of UAV multispectral data collection and BAWSeg dataset construction. The pipeline includes flight acquisition with RTK and irradiance sensors, radiometric calibration, orthorectification, stitching, feature extraction, annotation, and resulting crop--weed--soil maps.}
\label{fig:overview}
\end{figure}

To enable supervised learning and deployment-oriented evaluation in cereal systems, we introduce BAWSeg, a four-year UAV multispectral benchmark collected over two commercial barley paddocks near Kondinin, Western Australia (2020--2023), using a DJI Phantom~4 Multispectral platform with real-time kinematic (RTK) positioning and an upward irradiance sensor. Each campaign provides radiometrically calibrated five-band reflectance orthomosaics (blue, green, red, red edge, and near infrared), five derived vegetation indices, namely normalized difference vegetation index (NDVI), green normalized difference vegetation index (GNDVI), enhanced vegetation index (EVI), soil-adjusted vegetation index (SAVI), and modified soil-adjusted vegetation index (MSAVI)
, and dense pixel annotations for crop, weed, and other classes that explicitly address crop--weed canopy mixing. Unlike existing resources that are either close-range field imagery or aerial datasets with anomaly-style labels, BAWSeg is designed as a reproducible UAV orthomosaic benchmark with consistent radiometry and leakage-free spatial block splits. These splits support three evaluation protocols---within-plot, cross-plot, and cross-year (Figure~\ref{fig:overview})---so that generalization across paddocks and seasons can be measured under realistic domain shifts rather than optimistic random~sampling.

{
We evaluate VISA on BAWSeg under three deployment-oriented protocols, namely within-plot, cross-plot, and cross-year, so that performance can be examined under both in-domain testing and spatiotemporal transfer. The experimental study includes protocol-level evaluation, comparisons with representative baselines, and ablation analysis to assess whether decoupled radiance and vegetation-index modelling improves robustness across paddocks and seasons. Quantitative findings are presented in the Results Section and summarized in the Conclusions Section.
}

Our main contributions are summarized as follows:
\begin{itemize}
\item VISA 
 model: a two-stream network that integrates radiance and vegetation-index reasoning for improved segmentation under mixed canopy scenarios.
\item BAWSeg benchmark: a UAV multispectral dataset for barley fields with high-resolution, pixel-level crop, weed, and soil annotations under realistic field and seasonal variations.
\item Evaluation protocol: systematic within-plot, cross-plot, and cross-year benchmarks that demonstrate the utility of BAWSeg for model assessment and precision weed mapping research.
\end{itemize}

\section{Related Work}\label{sec:related}

We group prior work into three areas: UAV multispectral weed mapping with index-based pipelines, crop--weed semantic segmentation using convolutional encoder--decoder networks on RGB or multispectral inputs, and Transformer or state-space models for remote sensing segmentation. These studies provide the baselines and context for our two-stream radiance--index model.

%
UAV multispectral imagery is widely used for site-specific weed management and crop--weed discrimination, typically by deriving vegetation indices such as NDVI, GNDVI, SAVI and MSAVI from calibrated red, green, red edge and near-infrared bands and then classifying crop, weed and soil with threshold rules or shallow models including support vector machines, random forests and multilayer perceptrons~\cite{lopezgranados2011weedreview,pena2013maizeuav,lopezgranados2016sunflower,2025lidar,decastro2018rfobia}. Object-based image analysis combined with random forests has been used to segment homogeneous regions and generate prescription maps in maize and sunflower, while exploiting only local spectral and texture descriptors~\cite{castaldi2017maize,perezortiz2015semisupervised,ciceklidag2024high}. Semi-supervised systems and feature selection schemes further improve robustness under limited labels by stacking multiple vegetation indices into per-pixel or small-patch feature vectors for weed mapping in heterogeneous fields~\cite{perezortiz2016patterns,tamouridou2017mlpard,moshou2017novelty}. More recent work integrates multispectral UAV data with ancillary information and ensemble learners to estimate weed density at the plot scale in cereal crops~\cite{jin2023corn,2025citymultistream,ibrahim2025forest}. Across these studies, decisions are usually taken independently for each pixel or patch using hand-designed spectral indices and hard labels that assign a single class per pixel, which restricts the use of row-level and field-level structure and does not explicitly model the uncertainty of mixed pixels where crop plants and weeds occupy the same footprint.

%
UAV crop--weed mapping has increasingly been formulated as semantic segmentation with convolutional or encoder--decoder backbones such as U\mbox{-}Net, SegNet and DeepLabv3+, adapted from generic vision to field imagery \cite{ronneberger2015unet,badrinarayanan2017segnet,chen2018deeplabv3plus}. Early agricultural deployments trained fully convolutional networks on UAV or ground imagery to delineate crop rows and weed patches but largely followed single\mbox{-}stream designs \cite{lottes2017icra,fawakherji2023remotesensing,shahi2023drones}. Most studies use RGB or multispectral bands and append a small set of vegetation indices as extra channels, and then feed the concatenated tensor into one encoder that must learn spectral contrast and geometric texture without explicit separation of radiance and normalized index cues \cite{lottes2017icra,sa2018weedmap,fawakherji2023remotesensing}. Architectures that dedicate an index\mbox{-}aware branch or explicitly decouple spectral ratio information from spatial structure remain uncommon in the UAV literature, as summarized by recent reviews of deep learning for weed detection and drone imaging in agriculture \cite{li2023weedreview,remotesensing2023uavreview}. Supervision typically relies on hard one\mbox{-}hot pixel labels, so mixed pixels where crops and weeds co\mbox{-}occur are forced to a single class and label uncertainty is rarely modeled \cite{li2023weedreview,shahi2023drones}. These choices limit robustness under illumination change and can reduce sensitivity to small weed clusters embedded in crop canopies in \linebreak  orthomosaics \cite{sa2018weedmap,fawakherji2023remotesensing,olsen2019deepweeds}.

%
Transformer architectures have been adopted for remote sensing segmentation to capture long-range dependencies that are difficult for convolutional encoders. Early work adapts the Vision Transformer and Swin Transformer to aerial and satellite imagery using self-attention on patch tokens or shifted windows \cite{dosovitskiy2021vit,liu2021swin}, followed by hybrids such as SegFormer and related CNN Transformer models with lightweight decoders for high-resolution land cover from RGB or multispectral inputs \cite{xie2021segformer,zuo2022msvit}. Subsequent methods specialize Transformers for remote sensing through tailored tokenization and encoder--decoder designs, including RSTFormer and UNetFormer for urban scenes and river networks and SpectralFormer for hyperspectral land cover classification \cite{li2022rstformer,li2023unetformer,kong2022spectralformer}. {In agriculture, Transformer variants are mainly used for crop type mapping and field-level weed detection from UAV or satellite imagery with single-stream ViT-style backbones on reflectance stacks or RGB composites and hard pixel labels \cite{sandovalpillajo2025maizeweed,uysal2024weedsense}. Recent Mamba-based segmentation studies have expanded this direction in remote sensing. RS$^{3}$Mamba introduced a dual-branch visual state-space design for remote sensing image semantic segmentation, while MFMamba extended Mamba to multimodal remote sensing segmentation through image and DSM fusion \cite{ma2024rs3mamba,wang2024mfmamba}. Further 2024--2025 studies explored efficient or multiscale Mamba-based decoders for high-resolution scene parsing, including a semantic-Transformer-assisted real-time model, a Vision Mamba model with multiscale multi-frequency fusion, and LMVMamba with adaptation fine-tuning \cite{ding2024mamba_transformer,cao2025visionmamba,li2025lmvmamba}. Importantly, for agricultural scenes, EGCM-UNet addressed farmland remote sensing image semantic segmentation, and U-MoEMamba reported Mamba-based UAV segmentation results for cabbage heads under complex field conditions \cite{zheng2025egcmunet,li2025umoemamba}. Nevertheless, most Transformer- and Mamba-based approaches still employ a single feature stream without explicit vegetation-index reasoning, structured grouping modules such as Slot Attention, or annotation protocols that represent mixed crop--weed canopies and soft label uncertainty.
}

\section{{Materials} and Methods}

This section specifies the complete workflow used to build the BAWSeg benchmark and develop the VISA segmentation model. It first describes the UAV acquisition setup, radiometric calibration and preprocessing steps that produce georeferenced multispectral orthomosaics, followed by the annotation procedure, patch extraction, and leakage-free block splits that support within-plot, cross-plot, and cross-year evaluation. It then details the two-stream architecture, feature fusion, training objective, and implementation settings used consistently across all experiments.

\subsection{Dataset Construction}\label{sec:dataset}

We first give details of the BAWSeg dataset, since the segmentation model is designed to exploit the information available in this dataset. BAWSeg is a multispectral dataset collected by a UAV from repeated field campaigns. It covers platform and radiometric calibration, mission geometry for uniform coverage, and preprocessing that rectifies, coregisters, orthorectifies, and mosaics all bands. It then outlines patch extraction, leakage-free splits, augmentation, and a dense polygon-based pixel annotation protocol with quality control. The process preserves traceability and consistent radiometry and geometry for reliable training and evaluation.

\subsubsection{Acquisition Platform and Configuration}

Image capture used a DJI Phantom~4 Multispectral (P4M) remotely piloted aircraft system (RPAS) (SZ DJI Technology Co., Ltd., Shenzhen, China) with six synchronized cameras, five narrowband monochrome sensors (blue, green, red, red edge and near-infrared (NIR)) and one RGB sensor. {The platform includes an integrated spectral sunlight sensor. In our workflow, the irradiance associated with each single-band image was obtained from the XMP metadata field \texttt{Irradiance} stored in the TIFF output of the aircraft. These per-image irradiance records were retained together with the image metadata, while an RTK module with TimeSync referenced geotags to the camera center.} The gimbal was fixed at nadir ($-90^\circ$). Exposure control operated in manual mode with fixed white balance and ISO for each mission, and shutter speed was set to avoid motion blur at the commanded groundspeed. Radiometric reference data were collected with a calibrated Lambertian reflectance panel using nadir frames acquired before and after each mission and whenever the solar zenith changed noticeably during flight, with the irradiance sensor enabled for all captures. Simple capture quality gates monitored shutter speed, saturation, irradiance validity and gimbal status, and any pass that violated these checks was repeated at the same altitude and speed. During acquisition, EXIF metadata recorded exposure parameters, RTK position, ellipsoidal height, gimbal yaw, pitch and roll, and irradiance flags. After each mission, these fields were parsed into a CSV file with frame identifiers, timestamps, platform pose and quality indicators. All data were stored in a mission-specific directory hierarchy indexed by date and site with separate locations for raw multispectral and RGB frames, metadata tables and integrity checksums, which preserves capture conditions for the radiometric normalization and orthomosaic generation described in the next subsection.

\subsubsection{Flight Mission Geometry and Coverage}

\begin{figure}[H]
\includegraphics[width=\linewidth]{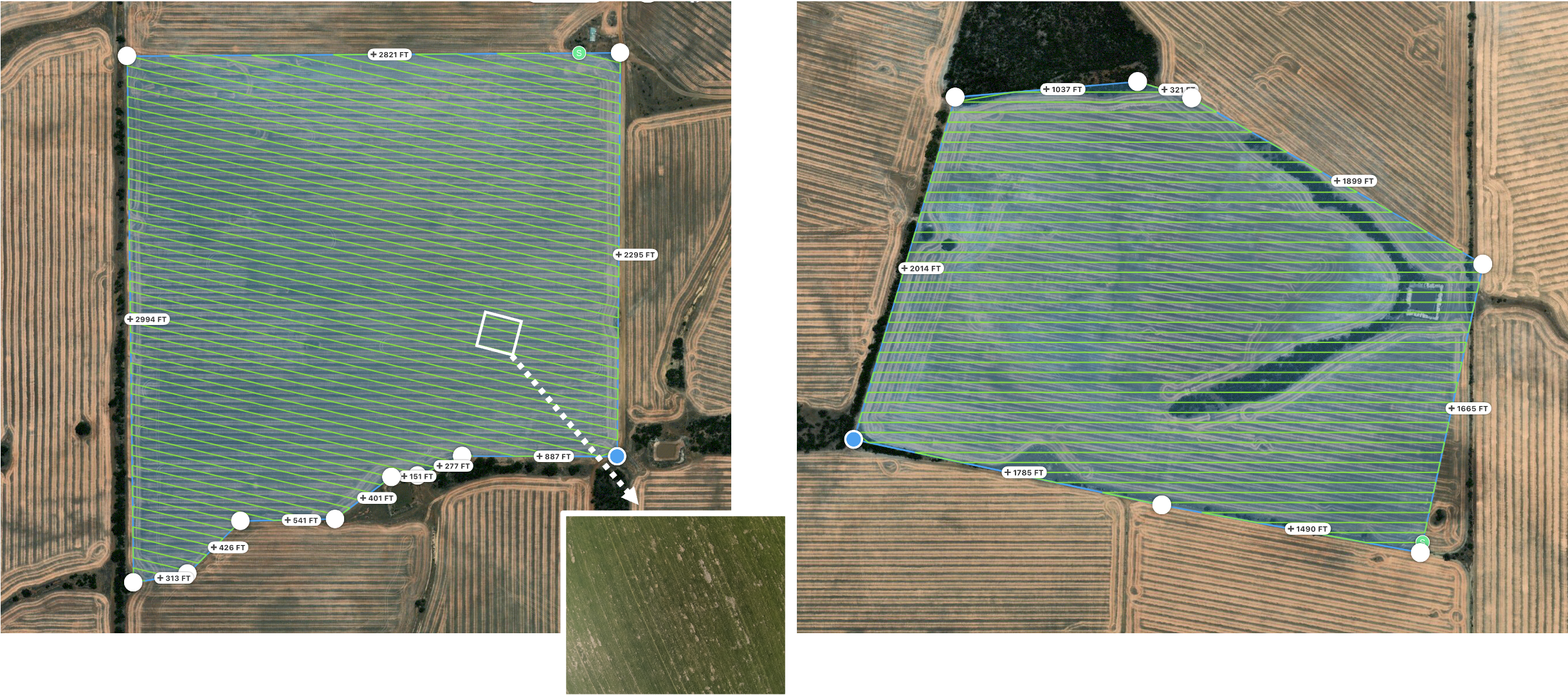}

\caption{{Planned} 
 Planned flight paths for the Kondinin experimental fields. Left, E8 polygon and survey lines planned in GSPro. Right, E2 polygon with the same overlap and altitude parameters. The parallel survey lines indicate the planned lawnmower flight trajectories within each field polygon. The inset shows a representative interior tile used for quality control of row orientation and texture.}
\label{fig:flightplan}

\end{figure}

Two experimental barley fields near Kondinin, Western Australia, were surveyed in four UAV campaigns from 2020 to 2023. The mapped polygons, centred at E2 (\text{lat }$-${32.508363,} 
 \text{lon }118.338139) and E8 (\text{lat }$-$32.516563, \text{lon }118.353799), cover \(0.6046\ \text{km}^2\) in total. Missions were planned in DJI Ground Station Pro (DJI GS Pro, version 2.0.17, SZ DJI Technology Co., Ltd., Shenzhen, China) using lawnmower patterns within each polygon, with the camera fixed at nadir, flight altitude set to \(120\ \text{m}\) above ground level, and both forward and side overlap set to \(80\%\). {For the DJI Phantom~4 Multispectral, this flight altitude corresponds to a nominal multispectral ground sample distance of approximately \(6.35\ \text{cm/pixel}\), following the manufacturer specification \((H/18.9)\ \text{cm/pixel}\), where \(H\) denotes the flight altitude relative to the mapped surface. At this resolution, crop-row structure and multi-row weed patches remain spatially resolved in the orthomosaic, which is consistent with the patch-level semantic segmentation setting used in BAWSeg. However, very small isolated weeds and heavily occluded seedlings may occupy only a few pixels, which limits separability at the finest spatial scale.} Line spacing, trigger intervals and groundspeed were derived from these parameters and applied uniformly during acquisition. Takeoff and landing points were placed on access tracks outside the cropped area and flights were scheduled between 10{:}30 and 14{:}30 local time, with segments repeated if wind or illumination changed noticeably to keep conditions consistent within each field. For every sortie, the mission plan and realised trajectory were logged with the field identifier, including polygon vertices, line spacing and commanded speed. Figure~\ref{fig:flightplan} illustrates the resulting coverage pattern, where survey lines follow the field boundaries and provide stable overlap for subsequent orthomosaic generation and feature matching across the four campaigns.

\subsubsection{Image Preprocessing}

\begin{figure}[H]
    \includegraphics[width=\linewidth]{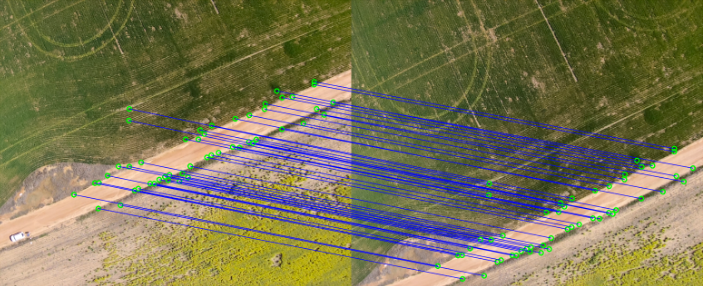}
    
    \caption{{Example} 
 of keypoint inliers for cross-band and inter-frame alignment. Green circles denote SIFT keypoints and blue lines denote matches retained after RANSAC.}
    \label{fig:sift}
    
\end{figure}

Raw multispectral frames were converted to single-precision arrays, filtered by capture quality flags, and denoised with a $3\times3$ median kernel applied independently to the R, G, B, NIR and RedEdge bands. {We selected this as a mild nonlinear denoising step to suppress isolated impulsive artifacts while preserving stronger local edges before cross-band registration and reflectance processing. However, because the operation is local, very small weed responses near the spatial resolution limit may be weakened when they occupy only one to two pixels.} Frames that contained more than 0.5\% saturated pixels in any band were discarded, and lens vignetting and distortion were corrected using the vendor calibration so that all subsequent steps operated on rectified images. Cross-band alignment then registered every band to the green reference. SIFT keypoints were extracted per band, descriptor matches were filtered with a Lowe ratio test and RANSAC, and a projective homography was estimated for each band, which is adequate for the near-planar scenes at 120 m altitude. The transform was refined by maximizing the enhanced correlation coefficient on a grid of small windows and registration quality is reported as the median reprojection error, with frames above a 0.5 pixel threshold flagged for reprocessing. Figure~\ref{fig:sift} illustrates typical inlier correspondences after matching and RANSAC.

Panel-based radiometric calibration converted the registered digital numbers to per-band reflectance that is consistent across sorties. {For each frame, the irradiance term was obtained from the per-image XMP metadata field \texttt{Irradiance} recorded by the aircraft. For each band, a scale factor was then derived jointly from the pre-flight and post-flight panel captures and these per-image irradiance values, and the frame reflectance was adjusted for exposure and illumination variation, clipped to a fixed range, and stored as \texttt{float32}.} Orthorectification then used RTK geotags and structure from motion in Agisoft Metashape Professional (version 2.0.1, Agisoft LLC, St. Petersburg, Russia) to estimate camera poses and a digital elevation model. All bands of each frame were treated as a rigid group and were projected onto the elevation model into a common ground coordinate system. Seamline optimization and blending within the overlap generated multispectral and RGB orthomosaics without exposure compensation for the analytical bands so that radiometric consistency was preserved. The final per-band mosaics were exported as tiled GeoTIFFs for index computation and patch extraction in the dataset pipeline.

\subsubsection{Annotation and Dataset Finalization}

\begin{figure}[H]
\includegraphics[width=\linewidth]{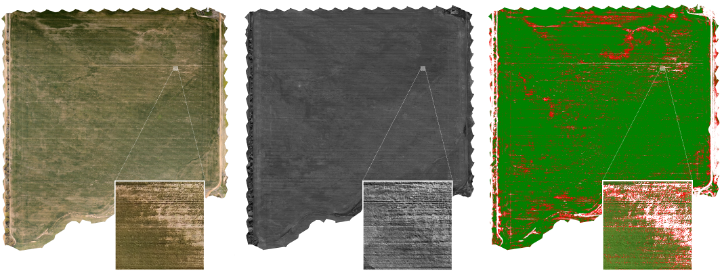}
\caption{{From} 
 left to right, RGB orthomosaic, a single-band reflectance map, and the hard label map. Green denotes crop, red denotes weed, and white denotes other.}
\label{fig:labelmaps}
\end{figure}

Pixels in the multispectral orthomosaics were labelled at native resolution as barley crop, ryegrass weed, or other, where other covers soil, wheel tracks, and infrastructure. Orthorectified and radiometrically normalized mosaics were tiled, and a vegetation mask derived from NDVI restricted annotation to canopy pixels while nonvegetation and invalid pixels were assigned other or an ignore value. Within this mask, annotators used QGIS (version 3.22 LTR, QGIS Development Team, Open Source Geospatial Foundation (OSGeo), Beaverton, OR, USA) with orthomosaic, NDVI and RGB composites to draw crop and weed polygons, and very small polygons were discarded. The polygons were rasterised by assigning each pixel the class of the polygon that contained its centre, overlaps were resolved in favour of weed polygons so that mixed canopies followed the weed label, and remaining vegetation pixels outside all polygons were kept as other. Each tile was annotated by one operator and checked by a second, and a stratified subset was reannotated to maintain consistent class usage across fields and years. The final labels were stored as single-channel GeoTIFFs with other equal to zero, crop equal to one, weed equal to two and ignore equal to two hundred and fifty five, and during training, this map was converted to one-hot targets \(y_{b,c,h,w}\in\{0,1\}\), with the ignore value forming a binary mask that removed those pixels from the supervised loss. Representative examples of the RGB orthomosaic, reflectance map, and final hard label map are shown in Figure~\ref{fig:labelmaps}.

\subsection{Methods}\label{sec:method}

\begin{figure}[H]
\includegraphics[width=\linewidth]{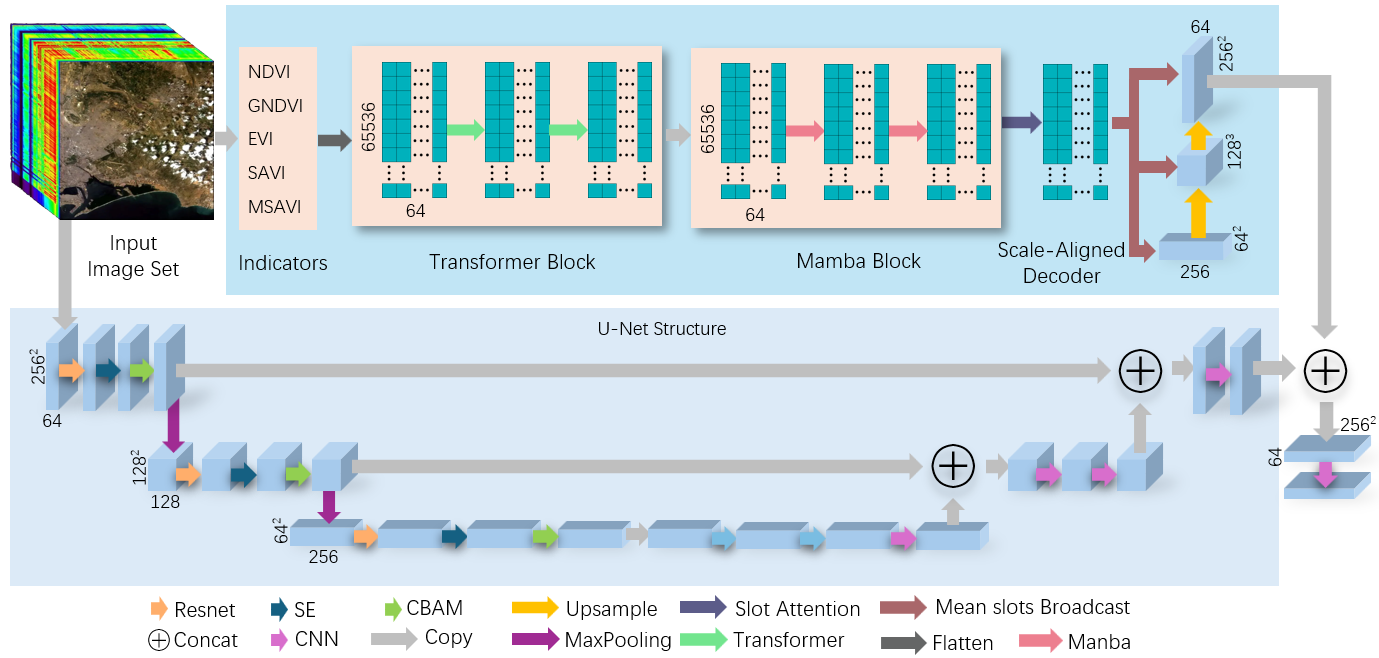}
\caption{
{Overview} 
 of the proposed pipeline. Bottom: SRAB radiance stream with residual attention and a U-Net decoder. Top: vegetation-index stream with windowed self-attention, Mamba blocks, Slot Attention with mean-slot broadcast, and a scale-aligned refinement module. Here, the scale-aligned refinement module denotes the lightweight feature-refinement stage that aligns the index-stream representation with the radiance branch on the native \(256\times256\) grid before fusion. Right: fusion head that concatenates the two 64-channel feature maps at the native resolution to generate class~logits.}
\label{fig:pipeline}
\end{figure}

We propose VISA, a two-stream segmentation architecture that leverages complementary cues from vegetation indices and multispectral imagery. The model aims to jointly capture fine spatial detail and field-scale agronomic patterns that are difficult to resolve from a single feature stream. As shown in Figure~\ref{fig:pipeline}, the network contains two parallel branches, a Vegetation Index Branch 
 that operates on normalized vegetation indices and a Spectral Residual Attention Branch (SRAB) that processes the multispectral radiance cube. Both branches output 64-channel feature maps at native \(256\times256\) resolution, which are fused by a shallow head to produce per-pixel logits for crop, weed, and background classes. This design separates low-level radiometric reasoning from high-level vegetation context modelling, improving robustness under illumination changes and canopy mixing. 
Let $B$ denote batch size. Each training sample is a $256\times256$ patch cropped from a georeferenced orthomosaic. The radiance input is a five-band reflectance tensor
$X^{\mathrm{raw}}\in\mathbb{R}^{B\times 5\times H\times W}$ with $H~{=}~W~{=}~256$, ordered as $(B,G,R,\mathrm{RE},\mathrm{NIR})$ after band co-registration.
The index input is a five-channel tensor $X^{\mathrm{idx}}\in\mathbb{R}^{B\times 5\times H\times W}$ formed by vegetation-index maps computed from the same reflectance bands.
Unless otherwise stated, convolutions use zero padding and preserve spatial size.
Both branches output native-resolution features in $\mathbb{R}^{B\times 64\times 256\times 256}$ and are fused to predict $C~{=}~3$ semantic classes $\{\text{other},\text{crop},\text{weed}\}$, while the ignore label is excluded from supervision by a binary mask.
The configuration used throughout the paper fixes the embedding width at $d~{=}~64$, the window size at $s~{=}~8$ for windowed self-attention, the number of Slot Attention groups at $K~{=}~6$, and the number of Slot Attention refinement iterations at $T~{=}~3$.

\subsubsection{Vegetation-Index Modelling Branch}\label{sec:vimb}

{
To complement raw-band learning with normalized spectral reasoning, we introduce a vegetation-index modelling branch that operates on a compact stack of five indices $\{\mathrm{NDVI}, \mathrm{GNDVI},  \mathrm{EVI}, \mathrm{SAVI}, \mathrm{MSAVI}\}$ derived from calibrated reflectance. These five indices were selected to provide complementary responses within the five-band sensing capability of the DJI Phantom~4 Multispectral. NDVI serves as a standard vegetation-vigor baseline. GNDVI increases sensitivity to chlorophyll variation through the green band. EVI incorporates the blue band and is less sensitive to canopy background and saturation at higher vegetation density. SAVI and MSAVI explicitly reduce soil-background effects, which is relevant in BAWSeg because the four-year barley scenes include both partially exposed soil and incomplete canopy closure. Therefore, we selected indices that jointly capture canopy vigor, chlorophyll-sensitive variation, reduced saturation, and soil-adjusted vegetation response, rather than relying on a single family of closely related ratios. Indices are computed from calibrated reflectance with a small stabilizer $\epsilon~{=}~10^{-6}$.
}
For example, $\mathrm{NDVI}=(\mathrm{NIR}-R)/(\mathrm{NIR}+R+\epsilon)$, $\mathrm{GNDVI}=(\mathrm{NIR}-G)/(\mathrm{NIR}+G+\epsilon)$, and $\mathrm{SAVI}=1.5(\mathrm{NIR}-R)/(\mathrm{NIR}+R+0.5+\epsilon)$.
We apply a fixed per-index standardization using statistics computed on the training split only.
For the $i$th index channel, $\tilde X^{\mathrm{idx}}_i=(X^{\mathrm{idx}}_i-\mu_i)/(\sigma_i+\epsilon)$, and the same $(\mu_i,\sigma_i)$ are reused for validation and test data to avoid leakage.

Each index map is linearly standardized and projected to an embedding width $d~{=}~64$ via a $1~{\times}~1$ convolution and LayerNorm{:}

\begin{equation}
Z = \text{LN}(W^{\mathrm{proj}} \tilde X^{\mathrm{idx}} + b^{\mathrm{proj}}), \quad
Z \in \mathbb{R}^{B\times d\times256\times256}.
\end{equation}
{Here,} 
 $W^{\mathrm{proj}}\in\mathbb{R}^{d\times 5\times 1\times 1}$ and $b^{\mathrm{proj}}\in\mathbb{R}^{d}$ implement a $1\times1$ projection from five indices to $d~{=}~64$ channels.
$\text{LN}(\cdot)$ denotes LayerNorm applied over the channel dimension at each spatial location, which stabilizes training under index magnitude variations across seasons and fields.

We reshape $Z$ into non-overlapping $8~{\times}~8$ windows and apply two lightweight Transformer encoder layers with windowed self-attention. This design preserves local spatial coherence (e.g., crop rows and patch textures) while avoiding the quadratic complexity of global attention. A decomposed relative positional bias further enhances geometric sensitivity within each window.

Concretely, $Z$ is partitioned into $N_w=(H/s)(W/s)$ non-overlapping windows.
For each window, we flatten its $s\times s$ grid into a token matrix $U\in\mathbb{R}^{n\times d}$ with $n=s^2$.
Multi-head attention uses $h$ heads with per-head width $d_h=d/h$.
For head $j$, queries, keys, and values are
$Q_j=U W^{Q}_j$, $K_j=U W^{K}_j$, and $V_j=U W^{V}_j$, where $W^{Q}_j,W^{K}_j,W^{V}_j\in\mathbb{R}^{d\times d_h}$.
The attention output is computed as
\begin{equation}
\mathrm{Attn}_j(U)=\mathrm{Softmax}\!\left(\frac{Q_j K_j^{\top}}{\sqrt{d_h}} + B^{\mathrm{rel}}_j\right)V_j,
\end{equation}
where $B^{\mathrm{rel}}_j\in\mathbb{R}^{n\times n}$ is a learnable relative positional bias indexed by 2D offsets within the window.
The $h$ head outputs are concatenated and projected back to width $d$ by $W^{O}\in\mathbb{R}^{d\times d}$.

Each encoder layer follows a pre-norm residual form
\begin{equation}
\hat U = U + \mathrm{WSA}(\mathrm{LN}(U)),\qquad
U^{+} = \hat U + \mathrm{FFN}(\mathrm{LN}(\hat U)),
\end{equation}
where $\mathrm{WSA}$ denotes the multi-head windowed attention above.
$\mathrm{FFN}$ is a position-wise two-layer perceptron with GELU nonlinearity applied independently to each token.
In our configuration, $s~{=}~8$ and $h~{=}~8$, which keeps attention local to $64$ tokens per window while preserving row-level texture cues.

To capture field-wide regularities, we append two Mamba-style state-space blocks that act as gated 1D filters along the token sequence. For token embedding $u_t$, the selective update follows:

\begin{equation}
x_{t+1} = \bar A \odot x_t + \bar B \odot (u_t \odot \sigma(W_g u_t))
\end{equation}
{{We} 
 apply the state-space blocks on a 1D token sequence obtained by a row-major raster scan of the $H\times W$ grid, giving $L~{=}~HW$ tokens. The current implementation uses a single scan order only and does not adopt bidirectional, reverse, or cross-scanning variants. At step $t$, $u_t\in\mathbb{R}^{d}$ is the token embedding and $x_t\in\mathbb{R}^{d}$ is the state vector. $\bar A\in\mathbb{R}^{d}$ and $\bar B\in\mathbb{R}^{d}$ are learned per-channel coefficients after discretization, and $\odot$ denotes element-wise multiplication. The gate $\sigma(W_g u_t)$ uses $W_g\in\mathbb{R}^{d\times d}$ and modulates input injection to suppress ambiguous mixed pixels. In this design, directional bias is partially mitigated because the sequential module operates after windowed self-attention with relative positional bias, while spatial augmentation by rotations and flips is applied during training. This recurrence updates each token in linear time in $L$ and does not incur the quadratic cost of global attention, which is important for field-scale context modelling on dense orthomosaics.
}


Finally, we apply Slot Attention ($K~{=}~6$ slots, $T~{=}~3$ iterations) to group tokens into structured semantic regions (e.g., crop canopy, weed clusters, soil). The aggregated descriptor is broadcasted back to the spatial grid, forming a feature map $F^{\mathrm{idx}}\in\mathbb{R}^{B\times64\times256\times256}$ that carries globally consistent vegetation cues. Let $U^{+}\in\mathbb{R}^{L\times d}$ denote the token matrix after windowed attention and state-space filtering.
Slot Attention maintains $K$ slot vectors $S=\{s_k\}_{k=1}^{K}$ with $s_k\in\mathbb{R}^{d}$.
At each refinement iteration, we compute attention weights between tokens and slots by
\begin{equation}
a_{t,k}=\frac{\exp\!\left(\langle W^{q}s_k,\;W^{k}u_t\rangle/\sqrt{d}\right)}
{\sum_{k'=1}^{K}\exp\!\left(\langle W^{q}s_{k'},\;W^{k}u_t\rangle/\sqrt{d}\right)},
\end{equation}
where $u_t$ is the $t$th token, $W^{q},W^{k}\in\mathbb{R}^{d\times d}$, and $\langle\cdot,\cdot\rangle$ is the dot product.
Slot updates aggregate values $v_t=W^{v}u_t$ with $W^{v}\in\mathbb{R}^{d\times d}$,
\begin{equation}
\Delta s_k = \sum_{t=1}^{L} a_{t,k}\,v_t,
\qquad
s_k \leftarrow \mathrm{GRU}(s_k,\Delta s_k) + \mathrm{MLP}(\mathrm{LN}(s_k)),
\end{equation}
where the GRU and MLP are shared across slots.
After $T$ iterations, we compute the mean-slot descriptor $m=\frac{1}{K}\sum_{k=1}^{K}s_k$ and broadcast it to all spatial locations by
$U^{b}_t = U^{+}_t + W^{b}m$ with $W^{b}\in\mathbb{R}^{d\times d}$.
This mean-slot broadcast injects a global grouping prior while retaining token-level variation for small weed clusters.
Finally, $U^{b}$ is reshaped back to $\mathbb{R}^{B\times d\times H\times W}$ and mapped to $F^{\mathrm{idx}}$ by a $3\times3$ convolution followed by normalization.

{
The resulting feature map is then passed to a scale-aligned refinement module, which consists of lightweight convolutional refinement layers applied on the native \(256\times256\) grid. Its role is to refine the index-stream representation before fusion and to align it with the radiance branch while preserving the contextual information aggregated by the Transformer, Mamba, and Slot Attention modules. The refined index feature map is additionally supervised by an auxiliary segmentation loss weighted by \(0.3\), which encourages discriminative learning in the index domain.
}

\subsubsection{Spectral Residual Attention Branch}\label{sec:srab}

{
In parallel, the spectral residual attention branch learns directly from calibrated multispectral reflectance $\{B, G, R, \mathrm{RE}, \mathrm{NIR}\}$ rather than from normalized vegetation-index ratios. This choice preserves local band-specific contrast at native resolution, including weak spectral discontinuities along crop--weed boundaries, fine row texture, and fragmented responses from small weed patches that may be attenuated after ratio-based compression. The branch adopts a three-stage residual-attention encoder followed by a symmetric decoder. Local residual convolutions capture short-range spatial detail, while the attention components reweight informative channels and spatial locations under task supervision.

Each stage stacks two \emph{residual-attention units} consisting of convolutional bodies, squeeze--excitation channel weighting, and CBAM-style spatial gating. Formally, a unit transforms feature $U$ as
}

\begin{equation}
Y = \mathrm{CBAM}\!\left(\mathrm{SE}\!\left(U + W_2 * \operatorname{GELU}(W_1 * U)\right)\right)
\end{equation}
{In} 
 this unit, $W_1$ and $W_2$ are $3\times3$ convolutions that preserve spatial size.
The residual branch $U + W_2 * \mathrm{GELU}(W_1 * U)$ increases nonlinearity while keeping gradient flow stable.

The squeeze--excitation operator produces channel weights from global statistics.
Let $g=\frac{1}{HW}\sum_{h,w}U_{:,h,w}\in\mathbb{R}^{C}$ be global average-pooled features for $C$ channels.
SE computes $a=\sigma\!\left(W_2^{\mathrm{se}}\mathrm{GELU}(W_1^{\mathrm{se}}g)\right)$ and rescales channels by $U^{\mathrm{se}}_{c,h,w}=a_c\,U_{c,h,w}$,
where $W_1^{\mathrm{se}}$ and $W_2^{\mathrm{se}}$ are learned linear layers and $\sigma(\cdot)$ is the sigmoid.

CBAM further applies spatial gating conditioned on local context.
Given $U^{\mathrm{se}}$, we form a spatial descriptor by concatenating average-pooled and max-pooled maps across channels and apply a $7\times7$ convolution to obtain a spatial mask $M^{\mathrm{sp}}\in[0,1]^{H\times W}$.
The output is $Y_{c,h,w}=M^{\mathrm{sp}}_{h,w}\,U^{\mathrm{se}}_{c,h,w}$, which suppresses background clutter and enhances thin weed structures aligned with crop rows.


{
Encoded features with widths $\{64,128,256\}$ are hierarchically aggregated by transposed convolutions and refinement blocks, restoring the native $256~{\times}~256$ resolution. The resulting feature map $F^{\mathrm{raw}}\!\in\!\mathbb{R}^{B\times64\times256\times256}$ encodes high-frequency appearance cues complementary to the vegetation-index pathway. Skip connections from the encoder to the decoder help preserve boundary localization and reduce the loss of thin or spatially sparse weed structures during downsampling.
}

The encoder uses three resolution levels with channel widths $\{64,128,256\}$.
Spatial resolution is reduced by a factor of two between levels, and the decoder restores the $256\times256$ resolution with learnable upsampling.
Skip connections from the three encoder levels are concatenated into the corresponding decoder stages to preserve row boundaries and small weed fragments that are lost under repeated downsampling.
The last decoder stage outputs $F^{\mathrm{raw}}$ at width $64$ to match the index branch for fusion.

Together, these two streams---one semantic and index-driven, the other spectral and residual-attentive---form a dual-branch encoder that captures both normalized vegetation signals and raw spectral variations for robust UAV weed segmentation.

\subsubsection{Feature Fusion and Prediction Head}

After decoding, the two streams provide aligned native-resolution feature maps with complementary content: the radiance stream preserves high-frequency textures and spectral contrast, whereas the index stream injects global semantic priors shaped by attention and state-space filtering. The fusion head performs local channel mixing and produces per-pixel logits for the target classes.

Both feature maps are aligned on the same $256\times256$ grid and share the same channel width, which allows fusion without interpolation.
The classifier predicts $C~{=}~3$ logits per pixel and the ignore label is implemented only in the supervision mask rather than as a prediction class.

Let $Y^{\mathrm{raw}},Y^{\mathrm{idx}}\in\mathbb{R}^{B\times 64\times 256\times 256}$ be the branch outputs at native resolution.
They are concatenated along the channel axis to form
$F^{\mathrm{cat}}=\mathrm{Concat}(Y^{\mathrm{raw}},Y^{\mathrm{idx}})\in\mathbb{R}^{B\times 128\times 256\times 256}$.
The fusion block first performs channel mixing and local aggregation by a $3~{\times}~3$ convolution with zero padding,
followed by batch normalization and a rectified linear activation. Denote the fusion kernel $W^{\mathrm{fus}}\in\mathbb{R}^{64\times 128\times 3\times 3}$ and bias $b^{\mathrm{fus}}\in\mathbb{R}^{64}$. The pre-activation fused map $F\in\mathbb{R}^{B\times 64\times 256\times 256}$ is defined component-wise as

\newcommand{\Wfus}{W^{\mathrm{fus}}}
\newcommand{\Bfus}{b^{\mathrm{fus}}}
\newcommand{\conv}{\mathop{\ast}}
\begin{equation}
F_{b,q,h,w} \;=\; (\Wfus \conv F^{\mathrm{cat}})_{b,q,h,w} \;+\; \Bfus_{q},
\qquad 1 \le q \le 64 .
\label{eq:fuse-compact}
\end{equation}
{Batch} 
 normalization with learned affine parameters $\gamma_q,\beta_q$ and an $\varepsilon$-stabilized variance is then applied per output channel, followed by a rectified activation. Writing $\mu_q$ and $\sigma_q^2$ for the batch-wise moments of $F_{b,q,:,:}$,

\begin{equation}
F'_{b,q,h,w}
=\max\!\left\{0,\;\gamma_q\frac{F_{b,q,h,w}-\mu_q}{\sqrt{\sigma_q^2+\varepsilon}}+\beta_q\right\}.
\end{equation}
The prediction head maps $F'$ to $C$ semantic classes by a $1{\times}1$ convolution $W^{\mathrm{cls}}\in\mathbb{R}^{C\times 64\times 1\times 1}$ with bias $b^{\mathrm{cls}}\in\mathbb{R}^{C}$ to produce the per-pixel logits $Z\in\mathbb{R}^{B\times C\times 256\times 256}$:

\begin{equation}
Z_{b,c,h,w}=\sum_{q=1}^{64}W^{\mathrm{cls}}_{c,q,1,1}\;F'_{b,q,h,w}+b^{\mathrm{cls}}_{c},
\quad 1\le c\le C.
\end{equation}

Class posteriors are obtained by a temperature-scaled softmax over the class dimension with $\tau=1$ during training and inference,
\begin{equation}
\begin{aligned}
P_{b,c,h,w}
&=\frac{\exp\!\big(Z_{b,c,h,w}/\tau\big)}
        {\sum_{c'=1}^{C}\exp\!\big(Z_{b,c',h,w}/\tau\big)},
P &\in \mathbb{R}^{B\times C\times 256\times 256}.
\end{aligned}
\label{eq:softmax}
\end{equation}
{Training} 
 minimizes a class-balanced cross-entropy on the fused logits with an ignore mask for unlabeled pixels. Let $\Omega$ be the set of labeled pixels over a mini-batch and $y_{b,c,h,w}\in\{0,1\}$ be the one-hot targets. Class weights $w_c$ follow median-frequency balancing with pixel frequencies $f_c$ measured on the training set, that {is,} 
 $w_c=\mathrm{median}(f_1,\ldots,f_C)/f_c$. The primary loss is
\begin{equation}
\mathcal{L}_{\mathrm{ce}}
=-\frac{1}{|\Omega|}\sum_{(b,h,w)\in\Omega}\ \sum_{c=1}^{C}
w_c\,y_{b,c,h,w}\,\log P_{b,c,h,w}.
\end{equation}
{To} 
 improve robustness to class imbalance and to sharpen thin weed boundaries, we add a soft Dice term and an edge-aware term computed on the labeled pixels.
Let $P_{b,c,h,w}$ be the fused posterior in~\eqref{eq:softmax} and $y_{b,c,h,w}$ be the one-hot target.
The Dice loss is
\begin{equation}
\mathcal{L}_{\mathrm{dice}}
=1-\frac{2\sum_{(b,h,w)\in\Omega}\sum_{c=1}^{C} P_{b,c,h,w}\,y_{b,c,h,w}+\epsilon}
{\sum_{(b,h,w)\in\Omega}\sum_{c=1}^{C} P_{b,c,h,w}+\sum_{(b,h,w)\in\Omega}\sum_{c=1}^{C} y_{b,c,h,w}+\epsilon}.
\end{equation}
{For} 
 edge supervision, we compute boundary maps by a fixed Sobel operator $\nabla$ applied channel-wise to the class posteriors and labels.
Define $E(P)=\sum_{c=1}^{C}\lVert\nabla P_c\rVert_1$ and \linebreak  $E(y)=\sum_{c=1}^{C}\lVert\nabla y_c\rVert_1$ on $\Omega$.
The edge loss uses an $\ell_1$ penalty
\begin{equation}
\mathcal{L}_{\mathrm{edge}}
=\frac{1}{|\Omega|}\sum_{(b,h,w)\in\Omega}\left|E(P)_{b,h,w}-E(y)_{b,h,w}\right|.
\end{equation}
{The} 
 fused objective is the weighted combination
\begin{equation}
\mathcal{L}_{\mathrm{fuse}}=\mathcal{L}_{\mathrm{ce}}+\lambda_{\mathrm{dice}}\mathcal{L}_{\mathrm{dice}}+\lambda_{\mathrm{edge}}\mathcal{L}_{\mathrm{edge}},
\end{equation}
where $\lambda_{\mathrm{dice}}$ and $\lambda_{\mathrm{edge}}$ are fixed across all experiments.

The total objective is $\mathcal{L}=\mathcal{L}_{\mathrm{fuse}}+\alpha\,\mathcal{L}^{\mathrm{idx}}_{\mathrm{aux}}$ with $\alpha=0.3$, where $\mathcal{L}^{\mathrm{idx}}_{\mathrm{aux}}$ is the auxiliary cross-entropy applied to $Y^{\mathrm{idx}}$ after a $1{\times}1$ prediction layer. Gradients flow through the fusion block into both branches. At inference, the per-pixel label is $\hat{c}_{b,h,w}=\arg\max_{c} P_{b,c,h,w}$ and the corresponding confidence is $\max_c P_{b,c,h,w}$, which is exported with the segmentation map for downstream decision modules.

\section{Results}

We report results for the proposed radiance-index two-stream model on BAWSeg under three deployment-oriented protocols, namely within-plot, cross-plot, and cross-year. Performance is computed on held-out spatial blocks using mean Intersection over Union (mIoU), per-class IoU, micro-averaged precision, recall and F1, overall accuracy (OA), and Cohen's $\kappa$ \cite{cohen1960agreement}. For mIoU confidence intervals (CIs), we use a block bootstrap that resamples test blocks with replacement and recomputes the metric per replicate \cite{efron1994bootstrap}. Unless stated otherwise, all metrics exclude ignore pixels.

\subsection{Experimental Setting}

\textbf{Hardware and software.}
All experiments ran on a single workstation with Ubuntu 22.04.4 LTS, an AMD Ryzen 9 7950X CPU, 32\,GB RAM, a 2\,TB NVMe SSD, and one NVIDIA GeForce RTX 4090 GPU with 24\,GB VRAM. The software stack used Python 3.10 and PyTorch 2.7 with CUDA 12.6. Training used automatic mixed precision, and evaluation used FP32. For reproducibility, we fixed a single global seed of 2026 across Python, NumPy, and PyTorch. We disabled cuDNN benchmarking and enabled deterministic algorithms where available, following PyTorch reproducibility guidance \cite{pytorch_randomness}.

\textbf{Training configuration.}
Models were trained for 50 epochs on $256\times256$ patches. Optimization used AdamW \cite{loshchilov2019decoupled} with $\beta_1~{=}~0.9$, $\beta_2~{=}~0.999$, weight decay $0.01$, and gradient clipping with a global norm of $1.0$. The initial learning rate was set to $6~{\times}~10^{-4}$ and decayed with a cosine schedule after a linear warm-up of 1500 iterations. The effective batch size was 16. Inputs were standardized per band or per index using training-set statistics. Data augmentation used spatial transforms only, including random horizontal and vertical flips and random rotations in $\{0^\circ,90^\circ,180^\circ,270^\circ\}$, which preserves calibrated reflectance semantics.

\textbf{Losses and inference.}
Training minimized a class-balanced cross-entropy on fused logits with an ignore mask, plus an auxiliary cross-entropy applied to the index stream with weight $\alpha~{=}~0.3$. Inference used single-scale evaluation at native patch resolution. For block-level evaluation, we applied a sliding window over each held-out block using $256\times256$ windows with stride 128 and averaged logits in overlapping regions before taking the argmax label map. This prevents boundary artifacts from dominating block~metrics.

\textbf{Metrics and confidence intervals.}
For class $c$, IoU is $\mathrm{IoU}_c = \frac{\mathrm{TP}_c}{\mathrm{TP}_c+\mathrm{FP}_c+\mathrm{FN}_c}$ computed over labeled pixels. Mean IoU averages over the three classes. Micro-precision, micro-recall, and micro-F1 are computed from the summed confusion counts across all classes and all blocks. Cohen's $\kappa$ is computed from the confusion matrix using standard chance correction for nominal labels \cite{cohen1960agreement}. For each reported CI, we draw 10{,}000 bootstrap replicates at the block level and take the 2.5 and 97.5 percentiles of the resulting mIoU distribution \cite{efron1994bootstrap}. Block-level resampling is used because adjacent pixels inside an orthomosaic are spatially correlated, and evaluation based on spatially independent partitions is the standard practice in remote sensing segmentation benchmarks \cite{wang2021loveda}.

\subsection{Segmentation Across Plots and Years}

\begin{figure}[H]
  \includegraphics[width=\linewidth]{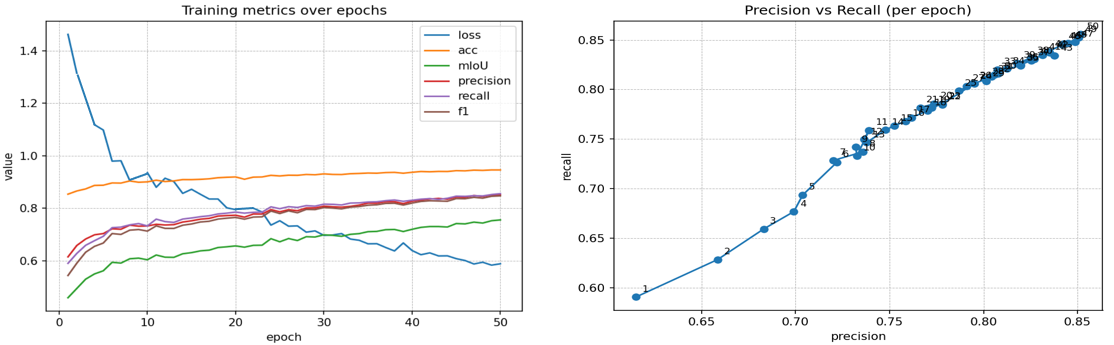}
  \caption{{Training} 
 dynamics of the proposed model. The upper panel shows loss, accuracy, mean IoU, precision, recall and F1 over 50 epochs. The lower panel shows precision versus recall for each epoch. Markers move towards the upper right region as training progresses and stabilise near the final operating point.}
  \label{fig:train_dynamics}
  
\end{figure}

Figure~\ref{fig:train_dynamics} summarizes training dynamics across 50 epochs. The fused loss decreases steadily and validation mIoU increases until convergence, with the best checkpoint selected by validation mIoU. The final checkpoint is used for all protocol evaluations.

Table~\ref{tab:peryear_within_full} reports within-plot performance stratified by year and field. Mean IoU remains in the range $[0.748, 0.763]$ across the eight year--field combinations, with weed IoU consistently above 0.62. Micro-F1 remains close to 0.84, while OA remains close to 0.95 and $\kappa$ remains close to 0.79. These results indicate stable behavior across acquisition years for the same operational region and sensor configuration.

\begin{table}[H]
\small 
\renewcommand{\arraystretch}{1.08}
\setlength{\tabcolsep}{3pt}

\caption{Within-plot segmentation performance per year and field on held-out test blocks. Mean IoU is estimated using $95\%$ bootstrap confidence intervals. Precision, recall and F1 are micro-averaged over all classes.}
\label{tab:peryear_within_full}

\begin{tabular*}{\columnwidth}{@{\extracolsep{\fill}}ccccccccccc}
\toprule
\multirow{2.5}{*}{\textbf{Year}} & \multirow{2.5}{*}{\textbf{Field}} &
\multirow{2.5}{*}{\textbf{mIoU}} &
\multicolumn{3}{c}{\textbf{Per Class IoU}} &
\multicolumn{3}{c}{\textbf{Micro Metrics}} &
\multicolumn{2}{c}{\textbf{Global Metrics}} \\
\cmidrule(lr){4-11}
 &  &  & \textbf{Crop} & \textbf{Weed} & \textbf{Other} & \boldmath{\textbf{$P$}} & \boldmath{\textbf{$R$}} & \textbf{F1} & \textbf{OA} & \boldmath{\textbf{$\kappa$}} \\
\midrule
\multirow{2}{*}{2020}
 & E2 & 0.748 & 0.842 & 0.621 & 0.781 & 0.838 & 0.846 & 0.842 & 0.943 & 0.784 \\
 & E8 & 0.751 & 0.844 & 0.623 & 0.786 & 0.840 & 0.848 & 0.844 & 0.944 & 0.786 \\
\midrule
\multirow{2}{*}{2021}
 & E2 & 0.754 & 0.845 & 0.626 & 0.791 & 0.842 & 0.850 & 0.846 & 0.945 & 0.789 \\
 & E8 & 0.759 & 0.848 & 0.630 & 0.800 & 0.844 & 0.852 & 0.848 & 0.946 & 0.791 \\
\midrule
\multirow{2}{*}{2022}
 & E2 & 0.762 & 0.850 & 0.636 & 0.801 & 0.846 & 0.854 & 0.850 & 0.947 & 0.795 \\
 & E8 & 0.760 & 0.849 & 0.633 & 0.798 & 0.845 & 0.853 & 0.849 & 0.947 & 0.793 \\
\midrule
\multirow{2}{*}{2023}
 & E2 & 0.757 & 0.846 & 0.632 & 0.794 & 0.843 & 0.851 & 0.847 & 0.946 & 0.792 \\
 & E8 & 0.763 & 0.851 & 0.638 & 0.800 & 0.848 & 0.856 & 0.852 & 0.948 & 0.799 \\
\bottomrule
\end{tabular*}
\end{table}

Table~\ref{tab:protocols_full} summarizes protocol-level generalization. Under within-plot evaluation, the model reaches mIoU $0.756 \pm 0.004$ with weed IoU 0.635. Under cross-plot transfer, training on one field and testing on the other yields mIoU near 0.71, and the primary degradation appears in the weed class, while crop IoU remains close to the within-plot level. Under cross-year testing, training on 2020--2022 and testing on 2023 yields mIoU $0.692 \pm 0.007$ and weed IoU 0.544, with crop IoU remaining above 0.83. {The larger drop in the cross-year setting is mainly attributable to temporal domain shift across acquisition years, which changes weed morphology, weed density, canopy mixing, exposed-soil proportion, and residual illumination conditions in the orthomosaics. This effect is concentrated in the weed class, while crop IoU remains comparatively stable, indicating that the degradation is driven primarily by year-dependent variation in weed appearance rather than by a general failure of preprocessing or radiometric calibration.} Across protocols, the relative sensitivity is concentrated in weed IoU, while OA stays above 0.93, reflecting the strong class imbalance between vegetation and the less frequent weed pixels.

\begin{table}[H]
\renewcommand{\arraystretch}{1.2}  
\caption{{Segmentation} 
 performance under different evaluation protocols. Train and test sets are defined at the block level. Mean IoU is reported with $95\%$ confidence intervals.}
\label{tab:protocols_full}
\setlength{\tabcolsep}{4.1pt}
\resizebox{\textwidth}{!}{%
\begin{tabular}{lccccccccccc}
\toprule
\multirow{2.5}{*}{\textbf{Protocol}} &
\multirow{2.5}{*}{\textbf{Train Set}} &
\multirow{2.5}{*}{\textbf{Test Set}} &
\multirow{2.5}{*}{\textbf{mIoU $\pm$ CI}} &
\multicolumn{3}{c}{\textbf{Per Class IoU}} &
\multicolumn{3}{c}{\textbf{Micro Metrics}} &
\multicolumn{2}{c}{\textbf{Global Metrics}} \\
\cmidrule(lr){5-12} 
 &  &  &  & \textbf{Crop} & \textbf{Weed} & \textbf{Other} & \boldmath{$P$} & \boldmath{$R$} & \textbf{F1} & \textbf{OA} & \boldmath{$\kappa$} \\
\midrule
Within plot &
E2 and E8, 2020 to 2023 &
E2 and E8, 2020 to 2023 &
$0.756 \pm 0.004$ &
0.847 & 0.635 & 0.794 &
0.851 & 0.843 & 0.847 &
0.946 & 0.794 \\
Cross plot &
E2, 2020 to 2023 &
E8, 2020 to 2023 &
$0.712 \pm 0.006$ &
0.840 & 0.576 & 0.724 &
0.838 & 0.830 & 0.834 &
0.937 & 0.768 \\
Cross plot &
E8, 2020 to 2023 &
E2, 2020 to 2023 &
$0.718 \pm 0.006$ &
0.846 & 0.584 & 0.724 &
0.842 & 0.836 & 0.839 &
0.939 & 0.773 \\
Cross year &
2020 to 2022, E2 and E8 &
2023, E2 and E8 &
$0.692 \pm 0.007$ &
0.832 & 0.544 & 0.700 &
0.828 & 0.816 & 0.822 &
0.936 & 0.752 \\
\bottomrule
\end{tabular}
}
\end{table}

\vspace{-6pt}

\subsection{Comparison with Existing Methods}

Table~\ref{tab:baseline_comparison} compares the proposed model with an index-based classical baseline and representative CNN and Transformer segmentation networks under the within-plot protocol. All deep baselines were trained and evaluated using the same block splits, patch size, optimizer type, learning-rate schedule family, and number of epochs as the proposed model. For multispectral inputs, we adapted the first layer of each architecture to accept five channels. For models that use patch embeddings, we replaced the input projection to accept five channels and initialized added channel weights by copying the mean of RGB weights when ImageNet pretraining was used, and then fine-tuned end-to-end on BAWSeg.

\begin{table}[H]
\renewcommand{\arraystretch}{1.08}  
\caption{{Comparison} 
 with classical baselines and modern segmentation networks under the within plot protocol on E2 and E8 from 2020 to 2023. MSI denotes the five calibrated multispectral bands and indices denote the five vegetation indices. Mean IoU is reported with $95\%$ bootstrap confidence intervals over test blocks.}

\label{tab:baseline_comparison}
\setlength{\tabcolsep}{2pt}
\resizebox{\textwidth}{!}{%
\begin{tabular}{@{}lcccccccccc@{}}
\toprule
\textbf{Method} & \textbf{Year/Venue} & \textbf{Input} & \textbf{mIoU} \boldmath{$\pm$} \textbf{CI} & \textbf{Crop IoU} & \textbf{Weed IoU} & \textbf{Others IoU} & \textbf{F1} & \textbf{OA} & \boldmath{$\kappa$} & \textbf{Params (M)} \\
\midrule
RF + Indices~\cite{breiman2001random}              
    & 2021 ML             & Indices       & 0.674 $\pm$ 0.010 & 0.812 & 0.532 & 0.679 & 0.734 & 0.907 & 0.703 & 0.2 \\
SegNet RGB~\cite{badrinarayanan2017segnet}                
    & 2019 TPAMI          & RGB           & 0.694 $\pm$ 0.009 & 0.825 & 0.558 & 0.699 & 0.752 & 0.915 & 0.718 & 29.4 \\
SegNet MSI~\cite{badrinarayanan2017segnet}                
    & 2024 TPAMI          & MSI           & 0.702 $\pm$ 0.009 & 0.831 & 0.567 & 0.708 & 0.759 & 0.918 & 0.724 & 29.5 \\
U-Net RGB~\cite{ronneberger2015unet}                 
    & 2023 MICCAI         & RGB           & 0.712 $\pm$ 0.008 & 0.836 & 0.579 & 0.720 & 0.768 & 0.922 & 0.732 & 31.0 \\
U-Net MSI~\cite{ronneberger2015unet}                 
    & 2022 MICCAI         & MSI           & 0.731 $\pm$ 0.007 & 0.842 & 0.594 & 0.758 & 0.783 & 0.930 & 0.748 & 31.1 \\
UNet++ MSI~\cite{zhou2018unetpp}                
    & 2024 DLMIA          & MSI           & 0.735 $\pm$ 0.007 & 0.844 & 0.603 & 0.759 & 0.787 & 0.932 & 0.751 & 34.2 \\
DeepLabv3 R50 RGB~\cite{chen2017deeplabv3}         
    & 2020 arXiv          & RGB           & 0.724 $\pm$ 0.008 & 0.838 & 0.585 & 0.748 & 0.775 & 0.928 & 0.743 & 41.1 \\
DeepLabv3+ R50 MSI~\cite{chen2018deeplabv3plus}        
    & 2024 ECCV           & MSI           & 0.738 $\pm$ 0.007 & 0.845 & 0.609 & 0.761 & 0.788 & 0.933 & 0.753 & 43.3 \\
PSPNet R50 RGB~\cite{zhao2017pspnet}            
    & 2023 CVPR           & RGB           & 0.720 $\pm$ 0.008 & 0.835 & 0.582 & 0.743 & 0.772 & 0.927 & 0.739 & 46.3 \\
HRNetV2 W18 MSI~\cite{wang2020hrnet}           
    & 2022 TPAMI          & MSI           & 0.740 $\pm$ 0.007 & 0.846 & 0.611 & 0.763 & 0.792 & 0.934 & 0.756 & 65.8 \\
BiSeNetV2 RGB~\cite{yu2021bisenetv2}             
    & 2021 IJCV           & RGB           & 0.707 $\pm$ 0.009 & 0.829 & 0.566 & 0.726 & 0.764 & 0.921 & 0.729 & 13.4 \\
SegFormer B0 RGB~\cite{xie2021segformer}          
    & 2021 NeurIPS        & RGB           & 0.732 $\pm$ 0.007 & 0.843 & 0.598 & 0.755 & 0.784 & 0.931 & 0.749 & 13.7 \\
SegFormer B1 RGB~\cite{xie2021segformer}          
    & 2021 NeurIPS        & RGB           & 0.736 $\pm$ 0.007 & 0.846 & 0.602 & 0.760 & 0.787 & 0.933 & 0.752 & 27.6 \\
SegFormer B1 MSI~\cite{xie2021segformer}          
    & 2023 NeurIPS        & MSI           & 0.744 $\pm$ 0.005 & 0.847 & 0.616 & 0.769 & 0.793 & 0.943 & 0.786 & 27.8 \\
Ours
    & 2026 & MSI
    & $0.756 \pm 0.004$
    & 0.847
    & 0.635
    & 0.794
    & 0.847
    & 0.946
    & 0.794
    & 22.8 \\
\bottomrule
\end{tabular}
}

\end{table}

The random forest baseline uses the five vegetation indices as input features and predicts per-pixel labels, followed by a $3\times3$ majority filter to remove isolated predictions. It achieves mIoU $0.674 \pm 0.010$ with weed IoU 0.532, which quantifies the gap between index-threshold style pipelines and learned dense segmentation on calibrated mosaics.

Among deep baselines, the strongest multispectral model in this comparison is SegFormer-B1 with MSI input, which achieves mIoU $0.744 \pm 0.005$ and weed IoU 0.616. Convolutional multispectral models such as U-Net MSI and DeepLabv3+ MSI achieve mIoU in the 0.73 to 0.74 range with weed IoU below 0.61. The proposed two-stream fusion achieves mIoU $0.756 \pm 0.004$ and weed IoU 0.635 with 22.8M parameters. Relative to SegFormer-B1 MSI, this corresponds to a +0.012 absolute improvement in mIoU and a +0.019 improvement in weed IoU under matched splits and evaluation. These gains are consistent with separating calibrated radiance cues from index cues and fusing them at native resolution, rather than forcing a single encoder to learn both signal types from a concatenated input.

\subsection{Ablation Study}

Table~\ref{tab:ablation} reports ablations under the within-plot protocol. Each variant modifies one factor while keeping the same training configuration, random seed, optimizer, schedule, and data splits. Reported mIoU values include block-bootstrap CIs computed with the same procedure as the main results. Throughput is measured at batch size 1 using FP16 inference on an RTX 4090, with 20 warm-up iterations and 200 timed iterations, and GPU timing synchronized before and after each iteration.

\begin{table}[H]
\renewcommand{\arraystretch}{1.08}  %
\caption{Ablation on the within‐plot protocol. $\Delta$ is relative to the full model. Params in millions. FLOPs in billions at $256\times256$. Mem in GB at FP16, batch size $1$. mIoU reports block-bootstrap 95\% confidence intervals. Arrows indicate whether higher or lower values are preferable.}
\label{tab:ablation}

\setlength{\tabcolsep}{5pt}
\resizebox{\linewidth}{!}{
\begin{tabular}{lcccccccc}
\toprule
\textbf{Variant} & \textbf{mIoU} {\boldmath{$\uparrow$}} 
 & \boldmath{$\Delta$} & \textbf{Weed IoU} \boldmath{$\uparrow$} & \textbf{Params~(M)} {\boldmath{$\downarrow$}} & \textbf{FLOPs~(G)} \boldmath{$\downarrow$} & \textbf{Mem~(GB)} \boldmath{$\downarrow$} & \textbf{FPS} \boldmath{$\uparrow$} \\
\midrule
\multicolumn{8}{c}{\emph{Module removal or replacement}} \\
\midrule
Full model & $0.756 \pm \CIMain$ & $0.000$ & 0.635 & \ParamCountM & 33.6 & 2.60 & 78 \\
w/o VIMB (RAW only) & $0.731 \pm \CIAbl$ & $-0.025$ & 0.596 & 9.6 & 25.4 & 2.20 & 92 \\
w/o relative bias in WSA & $0.749 \pm \CIAbl$ & $-0.007$ & 0.626 & 12.8 & 33.4 & 2.60 & 78 \\
w/o Slot Attention & $0.744 \pm \CIAbl$ & $-0.012$ & 0.618 & 12.3 & 32.4 & 2.50 & 80 \\
w/o mean-slot broadcast & $0.748 \pm \CIAbl$ & $-0.008$ & 0.622 & 12.3 & 32.4 & 2.50 & 80 \\
Single‐scale decoder (idx stream) & $0.747 \pm \CIAbl$ & $-0.009$ & 0.621 & 12.6 & 32.7 & 2.55 & 80 \\
\midrule
\multicolumn{8}{c}{\emph{Quantity changes}} \\
\midrule
WSA heads $h~{=}~4$  & $0.750 \pm \CIAbl$ & $-0.006$ & 0.627 & 12.8 & 33.5 & 2.60 & 78 \\
WSA heads $h~{=}~8$ (full) & $0.756 \pm \CIMain$ & $0.000$ & 0.635 & 12.8 & 33.6 & 2.60 & 78 \\
WSA heads $h~{=}~12$ & $0.758 \pm \CIAbl$ & $+0.002$ & 0.637 & 12.9 & 34.4 & 2.70 & 74 \\
Mamba layers $L~{=}~0$ & $0.738 \pm \CIAbl$ & $-0.018$ & 0.610 & 12.2 & 31.5 & 2.50 & 82 \\
Mamba layers $L~{=}~1$ & $0.746 \pm \CIAbl$ & $-0.010$ & 0.624 & 12.5 & 32.5 & 2.55 & 79 \\
Mamba layers $L~{=}~2$ (full) & $0.756 \pm \CIMain$ & $0.000$ & 0.635 & 12.8 & 33.6 & 2.60 & 78 \\
Mamba layers $L~{=}~3$ & $0.757 \pm \CIAbl$ & $+0.001$ & 0.637 & 13.1 & 34.9 & 2.70 & 75 \\
\bottomrule
\end{tabular}
}
\footnotesize
\emph{Notes.} FLOPs: measured with \texttt{fvcore} (count MACs as FLOPs), input $1~{\times}~C~{\times}~256~{\times}~256$, \texttt{eval()} and Dropout off. 
Mem: PyTorch max-allocated GPU memory (FP16, batch$=1$). 
FPS: average over 200 iterations after 20 warm-ups, timing wrapped by \texttt{torch.cuda.synchronize()}.
Unless noted, differences $\le\,$\CIAbl\ mIoU are not statistically significant under block-level bootstrap.
\end{table}

Removing the vegetation-index modelling branch yields the largest accuracy drop, reducing mIoU from $0.756$ to $0.731$ and weed IoU from 0.635 to 0.596, while reducing parameters and increasing FPS. This isolates the contribution of index-based context beyond what is learned from calibrated multispectral bands alone. Within the index stream, removing relative positional bias in windowed self-attention reduces mIoU to $0.749$, and removing Slot Attention reduces mIoU to $0.744$ with a concurrent weed IoU decrease, indicating that grouping tokens into region descriptors contributes to weed discrimination under canopy mixing. Removing the mean-slot broadcast reduces mIoU to $0.748$, which shows that re-injecting a global descriptor into the spatial grid affects final predictions beyond the slot aggregation itself.

Varying attention heads shows a compute--accuracy trade-off. Reducing heads to $h~{=}~4$ decreases mIoU to $0.750$ with a minor weed IoU decrease. Increasing heads to $h~{=}~12$ yields mIoU $0.758$ while reducing FPS due to higher attention cost. For the state-space component, removing Mamba layers produces a clear reduction to $0.738$ mIoU and 0.610 weed IoU, and performance improves monotonically up to two layers, with limited change from two to three layers at increased compute and memory. Across these variants, improvements below the ablation CI scale in Table~\ref{tab:ablation} should be interpreted as small under the block-bootstrap uncertainty, while changes above that scale correspond to consistent differences under the same evaluation procedure.

\section{Discussion}

The evaluation protocols expose two distinct regimes. When training and testing on disjoint spatial blocks from the same paddocks and seasons, VISA reaches 0.756 mIoU and 0.635 weed IoU, indicating that the model can separate weed patches from barley canopy at orthomosaic resolution while preserving row structure. Under cross-plot testing, mIoU drops to about 0.71 and the degradation concentrates on the weed class, while crop IoU remains close to the within-plot level. The same pattern holds under cross-year testing, where mIoU decreases to 0.692 and weed IoU to 0.544, with crop IoU still above 0.83. This behaviour is consistent with operational conditions in cereal systems, where crop appearance is relatively regular within a growth stage, while weed pressure, phenology, and within-row occlusion vary strongly across paddocks and seasons. Therefore, the results suggest that the remaining generalization gap is driven less by background and crop variation and more by shifts in weed morphology, density, and canopy mixing.

A key design choice in VISA is to prevent the radiance cues and index cues from competing inside a single encoder. Vegetation indices suppress many illumination and exposure effects by construction and encode cross-band interactions that correlate with chlorophyll and canopy vigor, which makes them informative for vegetation delineation across flights. At the same time, indices compress radiometric detail and can attenuate subtle within-canopy contrasts that differentiate sparse weeds from barley leaves, especially when the canopy is dense and when soil is largely occluded. Conversely, calibrated multispectral reflectance retains high-frequency textures, row boundaries, and small spectral deviations that are informative for fine weed clusters, but this signal is more sensitive to residual radiometric drift and to local mosaic artefacts. VISA keeps these cues in separate pathways so that the radiance stream can preserve boundary detail without being dominated by normalized ratios, and the index stream can capture stable vegetation patterns without being distracted by band-specific noise. The fusion at native resolution then combines boundary precision from the radiance stream with context from the index stream, which aligns with the observed gains over a single-stream SegFormer baseline at a similar scale.

The ablation study helps interpret which components contribute to weed robustness. Removing the index branch reduces mIoU by 0.025 and lowers weed IoU from 0.635 to 0.596, indicating that vegetation-index reasoning contributes materially beyond what can be learned from calibrated bands alone. Within the index branch, windowed self-attention supports local structure and row-consistent patterns, and the relative positional bias yields a consistent mIoU reduction when removed, which is expected because row geometry is expressed primarily through relative offsets rather than absolute location in a tile. The state-space layers provide the largest single improvement among the index-branch modules, with a 0.018 mIoU drop when they are removed, which suggests that linear-time context propagation helps encode plot-scale regularities that extend beyond the attention window. Slot Attention further improves weed sensitivity, with a 0.012 mIoU drop when removed, consistent with the role of region-level grouping in stabilizing representations of sparse and fragmented weed patches embedded in crop canopies. The mean-slot broadcast contributes an additional margin, indicating that injecting a global descriptor back to the spatial grid helps resolve local ambiguity in mixed pixels. 

{
The sequential component in the index branch also merits clarification with respect to scan direction. In the current model, the Mamba-style state-space blocks use a single row-major raster scan to convert the 2D feature grid into a 1D sequence. We do not use bidirectional or cross-scanning in this study. A multi-directional design could, in principle, reduce directional bias further for anisotropic field structure. However, in the present architecture, the sequential module is preceded by windowed self-attention with relative positional bias, and the training pipeline includes rotation and flip augmentation, which already reduce sensitivity to a fixed scan order. In addition, the inputs are orthorectified and radiometrically calibrated before sequence construction, so residual viewing and illumination effects are primarily handled in preprocessing rather than by changing scan direction. Under this setting, we did not observe evidence in the protocol-level results that scan order was a dominant source of error. Therefore, we retained the single-scan formulation to preserve the lightweight design of the index branch, while regarding multi-directional scanning as a possible extension for future work.
}

{
The selected index set should also be distinguished from the vegetation-index products provided in standard DJI software workflows. For Phantom~4 Multispectral data, DJI Terra provides NDVI, GNDVI, NDRE, LCI, and OSAVI, which only partially overlap with the index set used in this study. Our choice of NDVI, GNDVI, EVI, SAVI, and MSAVI was guided by the need to retain one blue-band index and two soil-adjusted indices that are directly relevant to mixed-canopy and exposed-soil conditions in BAWSeg. In future deployment-oriented settings, vendor-generated indices could be used as an alternative index source if one aims to simplify the preprocessing workflow. However, they would define a different input space from the one used here and, therefore, should be treated as a new variant that requires retraining and revalidation rather than as a direct replacement.
}

{
The platform also provides a co-registered RGB stream, which could be combined with multispectral data in future extensions. In such a setting, RGB would mainly contribute visible texture, edge, and morphological detail, while the multispectral bands provide red edge and near-infrared responses that are more informative for vegetation condition and crop--weed discrimination under mixed canopy and illumination variation. Their combination is, therefore, most relevant for boundary refinement, delineation of small weed patches, and visually ambiguous regions where appearance and spectral cues are both needed. We did not include RGB fusion in the present study because our objective was to establish a controlled benchmark for multispectral reflectance and vegetation-index modelling only. Adding RGB would define a different multimodal setting and should be evaluated as a separate extension under the same block-based protocols.
}

BAWSeg is constructed to make these conclusions meaningful under realistic deployment constraints. The dataset spans four seasons across two commercial paddocks, includes calibrated five-band orthomosaics and a consistent set of vegetation indices, and uses leakage-free spatial blocks for training and testing. Block splits are critical in orthomosaic learning because random sampling can place near-duplicate textures and lighting conditions in both training and test sets, yielding optimistic estimates that do not translate across paddocks. The cross-plot and cross-year protocols further stress the model under changes in weed distribution and season-specific appearance, which are the primary sources of failure in practice. The relatively stable crop IoU across protocols indicates that the calibration and preprocessing pipeline produces consistent crop appearance, while the weed IoU drop under shift highlights the need for modelling choices and training strategies that target rare, small, and occluded weed structures. {This limitation is related to both acquisition scale and preprocessing. At the nominal multispectral GSD of about \(6.35\ \text{cm/pixel}\) at 120 m altitude, BAWSeg is well suited to row-level and patch-level weed segmentation, but very small weeds near the spatial resolution limit are inherently more difficult to separate from crop canopy and mixed pixels. The \(3\times3\) median filter used in preprocessing improves robustness to isolated pixel noise and helps stabilize subsequent alignment and reflectance processing, but it can also attenuate one to two pixel weed responses before feature extraction. Therefore, we regard the current preprocessing as a practical trade-off that improves overall orthomosaic quality while leaving very small weed detectability as a residual limitation of the pipeline.}

{
From a precision agriculture perspective, the reported performance should be interpreted together with the efficiency results in Table~\ref{tab:ablation}. The full model uses 22.8~M parameters, 33.6 GFLOPs, 2.60 GB FP16 memory, and runs at 78 FPS for \(256\times256\) patches on a single RTX 4090. These results indicate that the model is practical for routine post-flight deployment on a standard workstation, where tiled orthomosaics can be segmented efficiently for same-cycle weed mapping, targeted scouting, and prescription preparation. In this workflow, the dominant time cost is more likely to come from orthomosaic generation and preprocessing rather than from network inference itself. The confidence maps produced by the softmax output further support risk-aware filtering, where low-confidence weed detections can be flagged for manual inspection instead of immediate treatment. This makes the current system more suitable for reliable offline decision support than for direct onboard UAV inference, which is consistent with common agronomic practice.
}

Several limitations remain. First, the geographic scope is limited to two paddocks in one grainbelt region and one UAV sensor configuration, so the reported cross-year robustness does not fully characterize transfer to different soil types, management practices, or imaging systems. Second, the labels represent three classes and resolve crop and weed overlap by polygon precedence, which is operationally convenient but does not capture the fractional composition of mixed pixels in dense canopy. This affects both training targets and evaluation, especially when weeds are small and partially occluded. Third, orthomosaic generation can introduce seamline artefacts and local blur that change with overlap and wind conditions, which can act as nuisance variation across campaigns. Addressing these constraints motivates several concrete directions. Expanding BAWSeg to additional farms and sensors would improve coverage of radiometric and agronomic variability. Introducing probabilistic or soft labels near ambiguous boundaries and in mixed-canopy regions would better match the underlying physical mixture. For generalization, domain generalization and test-time adaptation strategies that operate on calibrated reflectance and indices are promising, since the cross-plot and cross-year drops are concentrated in the weed class. Finally, weakly supervised extensions that exploit unlabeled mosaics from new seasons could reduce annotation burden while improving robustness to the shifts that are most salient in deployment.

\section{Conclusions}

In this paper, we presented BAWSeg, a four-year UAV multispectral benchmark for barley weed segmentation collected over two commercial paddocks in Western Australia. BAWSeg provides radiometrically calibrated five-band reflectance orthomosaics (blue, green, red, red edge, and near-infrared), derived vegetation-index maps, and dense pixel annotations for crop, weed, and other classes, together with leakage-free spatial block splits and deployment-oriented evaluation protocols. Building on this benchmark, we proposed VISA, a two-stream segmentation architecture that decouples radiance-based learning from vegetation-index reasoning and fuses both cues at native resolution to improve robustness under mixed-canopy conditions. Across the BAWSeg evaluation suite, VISA achieves a within-plot mIoU of 0.756 with a weed IoU of 0.635 and overall accuracy of 0.946 using 22.8~M parameters, and it retains an mIoU of 0.712 under cross-plot transfer and 0.692 under cross-year testing.

Beyond accuracy gains, our results show that explicitly modelling vegetation-index structure complements calibrated radiance features and improves weed sensitivity under challenging field variability. The ablation study supports this conclusion: removing the index branch yields a 0.025 mIoU decrease, indicating that index-driven context contributes materially to final predictions. BAWSeg and VISA together offer a reproducible foundation for benchmarking multispectral crop--weed segmentation under realistic spatial and temporal shifts, and for developing models that better generalize across seasons and paddocks. To facilitate follow-up research and practical adoption, we will release the BAWSeg dataset, the VISA implementation, and trained model weights upon publication.






\vspace{6pt}
\authorcontributions{
Conceptualization, H.W. and A.M.; Methodology, H.W., M.I. and A.M.; Software, H.W.; Validation, H.W., M.I., D.S. and X.W.; Formal analysis, H.W.; Investigation, H.W., M.I. and D.S.; Resources, M.I., D.S. and A.M.; Data curation, H.W., M.I., D.S. and X.W.; Writing---original draft preparation, H.W.; Writing---review and editing, H.W., M.I., D.S., X.W. and A.M.; Visualization, H.W. and X.W.; Supervision, A.M.; Project administration, A.M. and M.I.; Funding acquisition, A.M. and M.I. All authors have read and agreed to the published version of the manuscript.
}

\funding{
{This research received no external funding.} 
}

%

\dataavailability{A public preview subset of the underlying multispectral weed-detection dataset is available on IEEE DataPort: Haitian Wang, Xinyu Wang, Muhammad Ibrahim, Dustin Severtson, Ajmal Mian, ``Multispectral Remote Sensing for Weed Detection in West Australian Agricultural Lands'', IEEE DataPort, 11 February 2026, doi: \url{https://doi.org/10.21227/f8e1-5934}. The full BAWSeg benchmark dataset, together with the VISA implementation source code, trained model weights, and protocol files, will be released upon publication of this article.}

\acknowledgments{
{The} 
authors acknowledge the support from The University of Western Australia and the Department of Primary Industries and Regional Development (DPIRD), Government of Western Australia. We thank the field site owners and operational staff in the Kondinin region for enabling repeated UAV data collection across multiple seasons, and we acknowledge the assistance and advice from collaborators involved in field logistics, data curation, and annotation quality control. The authors reviewed and edited the content and take full responsibility for the final manuscript.
}

\conflictsofinterest{
The authors declare no conflicts of interest. 
}



\abbreviations{Abbreviations}{
The following abbreviations are used in this manuscript:
\\
\vspace{-16pt}
\noindent
\begin{longtable}[l]{@{}ll}

BAWSeg & Barley Weed Segmentation benchmark dataset\\
VISA & Vegetation-Index and Spectral Attention (two-stream segmentation network)\\
UAV & Unmanned Aerial Vehicle\\
RPAS & Remotely Piloted Aircraft System\\
RTK & Real-Time Kinematic\\
RGB & Red, Green, and Blue\\
NIR & Near-Infrared\\
RE & Red Edge\\
NDVI & Normalized Difference Vegetation Index\\
GNDVI & Green Normalized Difference Vegetation Index\\
EVI & Enhanced Vegetation Index\\
SAVI & Soil-Adjusted Vegetation Index\\
MSAVI & Modified Soil-Adjusted Vegetation Index\\
SIFT & Scale-Invariant Feature Transform\\
RANSAC & Random Sample Consensus\\
WSA & Windowed Self-Attention\\
SE & Squeeze-and-Excitation\\
CBAM & Convolutional Block Attention Module\\
GRU & Gated Recurrent Unit\\
mIoU & mean Intersection over Union\\
IoU & Intersection over Union\\
OA & Overall Accuracy\\
CI & Confidence Interval\\
FP16 & 16-bit Floating Point\\
FP32 & 32-bit Floating Point\\
\end{longtable}
}

\begin{adjustwidth}{-\extralength}{0cm}

\reftitle{{References}}

\bibliographystyle{Definitions/mdpi}

\PublishersNote{}
\end{adjustwidth}
\end{document}